\documentclass[runningheads]{llncs}
\usepackage{siunitx}
\usepackage{multirow}
\usepackage[T1]{fontenc}
\usepackage{xcolor,colortbl}
\usepackage{graphicx}

\usepackage[many]{tcolorbox}

\usepackage{adjustbox}

\usepackage{caption}

\newtcolorbox{custombox}[2][]{%
    enhanced,breakable,
    colframe=red!40!black, 
    colback=red!10, 
    arc=1mm,
    outer arc=1mm,
    fontupper=\small,
    fontlower=\small,
    coltitle=white!50,
    fonttitle=\bfseries,
    boxsep=1mm,
    left=1mm,
    right=1mm,
    top=1mm,
    bottom=1mm,
    before={\noindent},
    segmentation style={solid, red!40!black}, 
    title=#2,%
    #1
}

\usepackage{tikz}
\usetikzlibrary{shapes.geometric, arrows}

\tikzstyle{startstop} = [rectangle, rounded corners, minimum width=2cm, minimum height=1cm,text centered, draw=black, fill=red!20]
\tikzstyle{process} = [rectangle, minimum width=2.5cm, minimum height=1cm, text centered, draw=black, fill=blue!20]
\tikzstyle{decision} = [diamond, minimum width=3cm, minimum height=1cm, text centered, draw=black, fill=yellow!20]
\tikzstyle{arrow} = [thick,->,>=stealth]

\usepackage[ruled,vlined,linesnumbered,longend,algochapter]{algorithm2e}

\newenvironment{algorithm-custom}[1][htb]{
	
	\begin{algorithm}[#1]%
	}{
\end{algorithm}
}
\begin{document}
\title{Backtranslation and paraphrasing in the LLM era? Comparing data augmentation methods for emotion classification.}
%
%
\author{Łukasz Radliński \inst{1}\orcidID{0000-0002-7366-3847} \and
Mateusz Guściora\inst{1}\and
Jan Kocoń\inst{1}\orcidID{0000-0002-7665-6896}}
\authorrunning{Ł. Radliński et al.}
\titlerunning{Backtranslation and Paraphrasing in the LLM era?}
%
\institute{Department of Artificial Intelligence, Wrocław University of Science and Technology, Wyb. Wyspiańskiego 27, 50-370 Wrocław, Poland\\
\email{lukasz.radlinski@pwr.edu.pl}}
\maketitle              
\begin{abstract}
Numerous domain-specific machine learning tasks struggle with data scarcity and class imbalance. This paper systematically explores data augmentation methods for NLP, particularly through large language models like GPT. The purpose of this paper is to examine and evaluate whether traditional methods such as paraphrasing and backtranslation can leverage a new generation of models to achieve comparable performance to purely generative methods. Methods aimed at solving the problem of data scarcity and utilizing ChatGPT were chosen, as well as an exemplary dataset. We conducted a series of experiments comparing four different approaches to data augmentation in multiple experimental setups. We then evaluated the results both in terms of the quality of generated data and its impact on classification performance. The key findings indicate that backtranslation and paraphrasing can yield comparable or even better results than zero and a few-shot generation of examples.

\end{abstract}
\section{Introduction}
Improvements in the quality of the results of artificial intelligence (AI) systems in recent years have led to their more widespread application and increasing interest. Deep learning (DL), as the main engine of these changes, requires vast amounts of input data to produce good output. Thus, data collection and processing have become crucial. One branch of AI is Natural Language Processing (NLP), which deals with textual data, among others. DL models that handle such data require a large amount and quality of them. However, many domains still have a shortage of such data. 

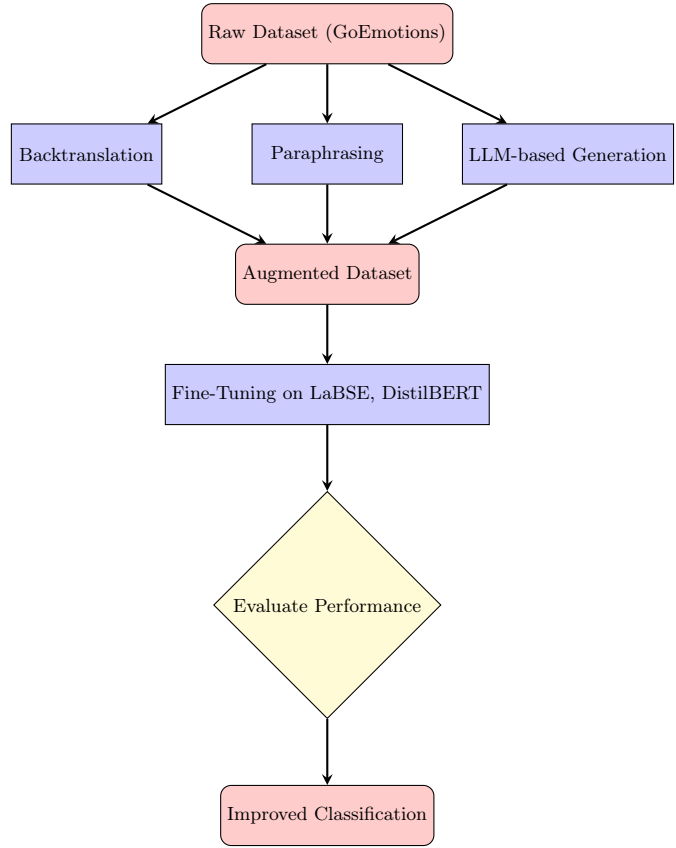
\begin{figure}[h]
    \centering
    \begin{tikzpicture}[node distance=2cm, scale=0.8, transform shape]

        \node (start) [startstop] {Raw Dataset (GoEmotions)};
        \node (backtrans) [process, below of=start, xshift=-4cm] {Backtranslation};
        \node (paraphrase) [process, below of=start] {Paraphrasing};
        \node (gen) [process, below of=start, xshift=4cm] {LLM-based Generation};

        \node (augdata) [startstop, below of=paraphrase, yshift=0cm] {Augmented Dataset};
        \node (train) [process, below of=augdata] {Fine-Tuning on LaBSE, DistilBERT};
        \node (eval) [decision, below of=train, yshift=-1.5cm] {Evaluate Performance};

        \node (result) [startstop, below of=eval, yshift=-1.5cm] {Improved Classification};

        \draw [arrow] (start) -- (backtrans);
        \draw [arrow] (start) -- (paraphrase);
        \draw [arrow] (start) -- (gen);

        \draw [arrow] (backtrans) -- (augdata);
        \draw [arrow] (paraphrase) -- (augdata);
        \draw [arrow] (gen) -- (augdata);

        \draw [arrow] (augdata) -- (train);
        \draw [arrow] (train) -- (eval);
        \draw [arrow] (eval) -- (result);

    \end{tikzpicture}
    \caption{Data augmentation for emotion classification}
    \label{fig:system_schema}
\end{figure}

Data augmentation (DA) is one technique for tackling this challenge. It has had a positive impact on computer vision and audio processing, and similar attempts are being made to augment text data. The manipulation of such data can vary in complexity and sophistication. Methods range from simple techniques like random insertion of characters to more advanced approaches, such as employing the generative power of large language models (LLMs) for paraphrasing entire data samples. LLMs, such as ChatGPT, are becoming more popular and accessible, making them valuable tools. Their generative capabilities enable the creation of extensive and coherent text samples. Although improvements in DL and NLP have been substantial, domain-specific tasks continue to face challenges due to data scarcity and the resulting class imbalance. The use of LLMs in NLP presents a promising new approach to DA. In this paper, we conducted a comprehensive comparison of multiple DA techniques on the GoEmotions classification dataset \cite{demszky2020goemotions}. We chose the most suitable categories to augment the dataset. Then we compared the conventional LLM generative approach to more traditional methods, i.e. backtranslation and paraphrasing, performed using language models. We then conducted a detailed analysis of the lexical diversity and semantic fidelity of the generated data. In order to verify the impact of the augmentation, we then used the augmented dataset to fine-tune two popular transformer models and compared the increase in performance across experimental setups, see Fig.~\ref{fig:system_schema}. The analysis of results gives key insights into how the new generation of language models can effectively leverage traditional approaches to data augmentation. All the code and prompts used to carry out the experiments are publicly available on GitHub  \footnote{\url{https://github.com/marentoo/data\_augmentation\_text\_classification\_task.git}}.

\section{Related Work}
Looking at a survey published last year on data augmentation \cite{zhou2024surveydataaugmentationlarge}, two main groups of data augmentation techniques are commonly used. The first group that saw a significant increase in popularity over the past years is the generated content-based approaches. Generative models have been utilized for data augmentation ever since they appeared. Works such as \cite{feng2020genaug} and  \cite{kumar2020data} showed that transformer models could effectively increase data diversity while preserving semantic fidelity. The works of \cite{liu2020data} have shown that such augmentation can directly increase classification performance. The release of GPT-3 cemented the role of generative models in data augmentation, as multiple studies have shown that they can be utilized to increase performance in classification tasks \cite{balkus2022improving}. Various techniques were used to increase the effectiveness of these models in data augmentation, such as fine-tuning the model\cite{zheng2022augesc} or using reinforcement learning \cite{liu2020data}. However, one of the most successfully employed techniques has been, without a doubt, Few Shot Learning (FSL). Numerous studies showed its effectiveness for models such as GPT-3 \cite{yoo2021gpt3mix} and ChatGPT \cite{moller2023prompt,kocon2023chatgpt,dai2023auggpt,koptyra2023clarin,kocon2023deep,wozniak2023big,kazienko2023human,wozniak2024personalized,kochanek2024improving,ferdinan2025fortifying,ngo2025integrating}. However, looking at the survey dedicated to text data augmentation with Large Language Models \cite{chai2025textdataaugmentationlarge}, one of the crucial challenges in using LLM is to ensure the diversity and quality of generated data.

The second, less popular, but still common group of methods listed in \cite{zhou2024surveydataaugmentationlarge} are label-based methods. These largely focus on using dataset labels or their embeddings to improve sampling quality or text generation. Although often slightly more complex than generation-based approaches, They have been proven to substantially improve the quality of generated data \cite{sun2020mixup,chen2020mixtext,wu2019conditional} and even allow for explainability in data augmentation \cite{kwon2023explainability}.

Another commonly employed technique for data augmentation that has not been listed in \cite{zhou2024surveydataaugmentationlarge} since it predates Large Language Models is backtranslation. The core idea to augment data by translating text into a foreign language and then back to the original has been used effectively long before generative devices such as BART or GPT were released \cite{fadaee2017data}. It was limited by the quality of the translation tools. However, only a few years later, models from the Workshop on Machine Translation (WMT models) \cite{xie2020unsupervised} and then the seq2seq transformers \cite{beddiar2021data} would show greater and greater improvements in the achievable results. Although the backtranslation technique predates the Large Language Models, it is still utilized \cite{xu2023} and was even effectively used to increase the quality of instructions for Large Language Models \cite{nguyen2024betteralignmentinstructionbackandforth,qi2024constraintbacktranslationimprovescomplex,li2024selfalignmentinstructionbacktranslation}.

\section{Dataset - GoEmotions}
\begin{figure}[!ht]
      \includegraphics[width=0.7\textwidth]{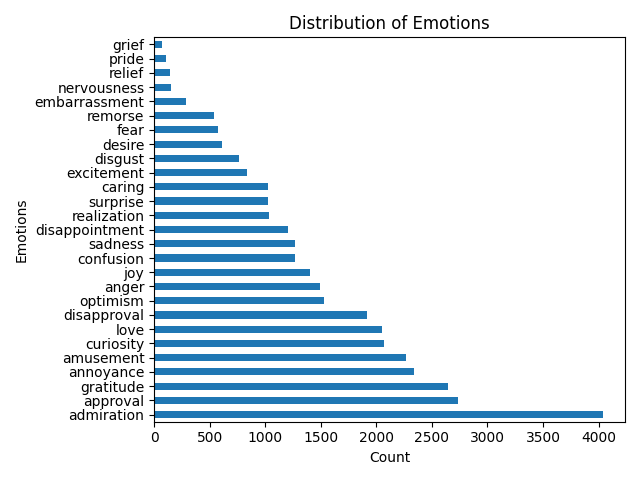}
      \caption{Emotion Labels Distribution}\label{fig:each_label_count_egoemo}
  \end{figure}

We decided to conduct our experiments on the GoEmotions dataset. This extensive dataset comprises textual comments labeled with specific emotions. Collaboratively developed by researchers from Google and Amazon, the dataset focuses on emotion-related tasks. It uses data from Reddit from 2005 to 2019, meticulously curated from subreddits with at least 10,000 comments, and exclusively in English \cite{demszky2020goemotions}. GoEmotions is a notably large dataset that benefits from human annotation, resulting in high ground truth accuracy. These attributes are particularly valuable given the scarcity of similar public datasets and the relative expense of human annotation. However, the data set has some obvious drawbacks that researchers point out as well. The dataset contains biases and is not representative of global diversity due to taking data specifically from Reddit. Notably, it is a multi-label dataset with 27 emotional labels. Therefore, in our research, we decided to focus on five labels that are least represented in the dataset. Looking at the class distribution of the labels, Figure~\ref{fig:each_label_count_egoemo}, there is a visible class imbalance in the dataset. It exhibits very limited representation in certain emotional categories. The five emotions with the least representation are:
 \begin{itemize}
     \item embarrassment (291 samples),
     \item nervousness (156 samples),
     \item relief (145 samples),
     \item pride (105 samples),
     \item grief (75 samples),
 \end{itemize}

  \begin{figure}[!ht]
      \includegraphics[width=0.7\textwidth]{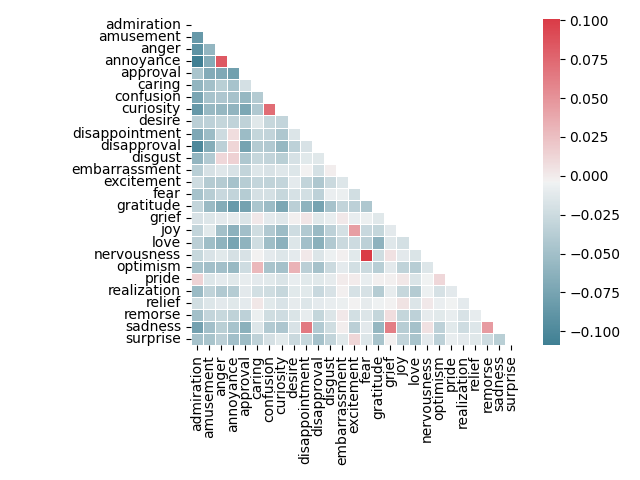} 
      \caption{Correlation Heatmap of labels}\label{fig:corellation_heatmap}
  \end{figure}
Furthermore, looking at the correlations between individual emotions, Figure~\ref{fig:corellation_heatmap}, there are visible dependencies between some of the labels. These include a high correlation between annoyance and anger, sadness and grief, remorse and sadness, disappointment and sadness, fear and nervousness, desire and optimism, excitement and joy, as well as confusion and curiosity. This means that many models may struggle to distinguish between these labels, even more so given the significant imbalance of the dataset. These qualities make the GoEmotions dataset very suitable for the data augmentation task. In our experimental setup, we generated additional samples for each of the five least represented classes and then used the modified data set to fine-tune the models used for evaluation.

\section{Data Augmentation Methods}
In our study, we wanted to focus on novel approaches to data augmentation using LLM. We decided to conduct our research using four data augmentation techniques:
\begin{itemize}
    \item Oversampling with dummy copies - baseline method
    \item Paraphrasing via prompting on a sentence level
    \item Generating via prompting with zero-shot learning and few-shot learning
    \item Backtranslation
\end{itemize}
Our goal is to verify both the quality of the generated data and its impact on the utility of the dataset.
\subsection{Baseline - Oversampling}
Oversampling is a method used to address the problem of class imbalance in machine learning. It
involves artificially manipulating the dataset to mitigate this issue. One of the most well-known techniques in this
context is the Synthetic Minority Oversampling Technique (SMOTE). Oversampling is used to
mitigate class imbalance by increasing the instances of the minority class. In its fundamental
form, it involves randomly duplicating instances of this class or classes. We use this approach as a baseline for the research as it provides no linguistic diversity while maintaining maximum semantic similarity to the original dataset. These two criteria are the main aspects of the quality of the generated data that we wanted to evaluate. Therefore, oversampling becomes the obvious choice as a baseline method. We compare two datasets augmented through oversampling. One is modified with three samples of each example in augmented classes, while the second is modified with five.

\subsection{Paraphrasing}
Paraphrases are texts that have been rewritten or rephrased using different words, structures, or forms. While the semantic meaning of a paraphrased sentence remains the same, its lexical and syntactic structures may differ. Paraphrasing is a highly effective and valuable method for data augmentation in NLP problems because it maintains semantic fidelity while introducing lexical diversity. Through such an enhancement, the model can learn new relationships and expand its vector space. By enhancing the scope of wording, the model can also prevent overfitting to specific linguistic patterns. This may have a wide range of applications, such as language understanding, summarization, generation, and translation \cite{beddiar2021data}. In our experiments, we utilize GPT-3.5 and GPT-4.0, which have already shown great potential for data augmentation \cite{moller2023prompt,dai2023auggpt}. Both models are used in three prompt configurations:
\begin{itemize}
    \item Prompt 1 - iteratively asking the model for a single paraphrase of each sample multiple times
    \item Prompt 2 - Nmax - asking models for generating multiple paraphrases at once for each sample and introducing as many as possible to the dataset
    \item Prompt 2 - Nbal - asking models for generating multiple paraphrases at once for each sample and introducing an equal number of samples from each class to the dataset
\end{itemize}

\subsection{Generating via Zero-shot and Few-shot generation}
Zero-shot learning (ZSL) and few-shot learning (FSL) are classes of methods used in machine learning to increase the generalizability of a model, Figure~\ref{tab:example_zsl_sentences_gpt-35}. These techniques have been widely adopted in classification problems where data is limited. The idea is to use either a few examples, in the case of FSL, or no examples at all, in the case of ZSL, to employ a trained deep learning model for a chosen new task. The task requires a model to infer classes that it has never seen during training (zero-shot learning) or has only seen a small number of times (few-shot learning). Methods designed to address data scarcity and class imbalance typically prioritize low-resource approaches. Zero-shot and few-shot learning are very resource effective. When applied using language models, these methods can be particularly effective as they also benefit from in-context learning. In terms of performance and results, FSL often yields better results than zero-shot, but it requires samples and more resources. FSL can achieve results comparable to those obtained through fine-tuning, and in certain tasks, these methods are even capable of outperforming human-level performance \cite{mann2020fsl,radford2019language,Gilardi_2023}. In our experiments, we used both GPT-3.5 and GPT-4 in zero-shot and 5-shot scenarios. This approach cannot be easily compared to backtranslation and paraphrasing in terms of data diversity and semantic fidelity. However, it is important to include it since it is arguably the simplest and most resource-effective LLM-based augmentation method.

\begin{figure}[ht]
\centering
\begin{custombox}{Example - Zero Shot Learning}
    \textbf{Prompt:} "Generate N different sentences in various forms that express a strong emotional sentiment for the following emotion:"\\[0.3em]
    \textbf{System message:} "You are a helpful assistant. Output sentences separated by newline in reply to <prompt>. Sentences should vary in type, slang, length, structure, tone and style, sentences such as comments, responses, opinions, and facts. It’s not necessary to often use the emotion’s name in every sentence. Do not number output or use bullet point for the output." \\[0.3em]
    \textbf{Example output:}
    ["I'm trying to laugh it off, but inside I'm dying of embarrassment.",
    "I can't believe I just did that!",
    "I still cringe thinking about that time I tripped and fell in front of everyone.",
    "Embarrassment can make you wish the ground would open up and swallow you whole.",
    "You know that feeling when you accidentally send a text meant for your friend to your boss? Yeah, pure embarrassment.",
    "Embarrassment can turn your face as red as a tomato."]
\end{custombox}
\caption{Example ZSL, input and output for GPT model}
\label{tab:example_zsl_sentences_gpt-35}
\end{figure}

\subsection{Backtranslation}
Backtranslation (BT) is a specific type of paraphrasing commonly used to generate semantically similar text with different lexical features. This process requires the source text to be translated into another language and then translated back into the source language. The paraphrases generated in this process are essentially by-products of the translation process by a model. Using various languages to generate new text content results in diversity, potentially enhancing model performance and expanding the dataset. The produced text exhibits diversity because different languages have unique structures, syntax, and linguistic features. Furthermore, imperfect translations can lead to numerous variations in the translated text \cite{fadaee2017data}. Researchers have tested different variations of BT. These variations include employing a single pair of languages, using multiple languages at different levels, and making architectural modifications to diversify the output \cite{abonizio2021,xie2020unsupervised}. With this method, certain challenges also arise. Often, translation models tend to favor high-frequency words, which can lead to repetition and limit diversity. This can result in overfitting the data and losing subtle details. Furthermore, the quality of the results is highly dependent on the quality of the translation model \cite{liu2020data,kwon2023explainability}.
Translation models tend to perform with various translation qualities depending on the target language. Using only one language for backtranslation also vastly limits the number of paraphrases generated. Therefore, to increase the number of models that can be compared to each other and reach a sufficient number of generated examples, we decided to perform backtranslation on multiple languages for each model and then aggregate all the examples into one dataset.  In this study, experiments involving backtranslation were conducted using the following models (translating into the following languages):
\begin{itemize}
    \item DeepL (Russian, Polish, Finnish, Japanese, Chinese, Bulgarian, Spanish, Hungarian, Greek, Turkish)
    \item GPT-3.5 (Polish, Chinese, Russian, Hindi, Hungarian, Finnish, Spanish, Japanese, Turkish, Arabic)
    \item GPT-4 (Polish, Chinese, Russian, Hindi, Hungarian, Finnish, Spanish, Japanese, Turkish, Arabic)
    \item GPT-4-turbo(Polish, Chinese, Russian, Hindi, Hungarian, Finnish, Spanish, Japanese, Turkish, Arabic)
    \item MarianMT model family (Hindi, Polish, Hungarian, Finnish, Russian, Chinese, Spanish, Japanese, Turkish, Arabic)
\end{itemize}
Most of the languages used for the backtranslation, but some of them differ due to a limited common group of sufficiently foreign languages to English available for all solutions.

\section{Results Evaluation}
We evaluate the increase in data in terms of both the quality of the generated data and the increase in classification performance achieved through augmentation. The evaluation of data quality is performed by measuring the linguistic diversity and the semantic fidelity achieved. The evaluation involved analyzing every fourth sentence pair between the reference and the generated text. The results were recorded for each pair, followed by calculating the average for the entire set. Such actions were repeated for every generated set. Assessment of the increase in classification is achieved by fine-tuning two reference models and comparing the results across the experimental setups. 

\subsection{Linguistic Diversity}
There are multiple known metrics that are used to measure lexical diversity. In order to receive more nuanced information, we decided to utilize four metrics for measuring diversity:
\begin{itemize}
    \item word count and word count ratio
    \item Jaccard dissimilarity
    \item information entropy
    \item ratio of the Type Token Ratios (TTR)
\end{itemize}

The word count ratio is calculated as the ratio of average word counts between original sentences and paraphrases.
Jaccard dissimilarity (\ref{eq:jaccard_dis}) is calculated as follows:
\begin{equation}
J_{\text{dissimilarity}}(A, B) = 1 - \frac{|A \cap B|}{|A \cup B|}
\label{eq:jaccard_dis}
\end{equation}

where A and B are two sentences to compare. These are reference sentences and generated sentences.  Information entropy is calculated as the ratio between the information entropy  (Figure~\ref{eq:information_entropy}) of the generated and reference sentences. The calculation of the information entropy for one of these sentences is as follows:

\begin{equation} H(S) = -\sum_{i=1}^{n} P(w_i) \log_2 P(w_i)
\label{eq:information_entropy}
\end{equation}

TTR ratio is calculated as the ratio between TTR (\ref{eq:ttr_ratio}) of the generated set of sentences and divided by the original. Each TTR is calculated as follows.
\begin{equation}
TTR = \frac{\text{types}}{\text{tokens}}
\label{eq:ttr_ratio}
\end{equation}

\subsection{Semantic fidelity}

Many techniques and studies emphasize measuring semantic fidelity, which is the similarity in meaning between the new text and the original. For data augmentation and, consequently, for subsequent machine learning tasks, maintaining semantic fidelity is crucial. Preserving the meaning and the label is key to determining the effectiveness of a model. There are two most common approaches to measuring semantic fidelity. The first is to use distance measures such as cosine similarity, dot product similarity, or Euclidean distance to calculate the distance between vector representations of the latent space of text. The other is BERTscore, which employs BERT embeddings to grasp semantic relations between texts. In our experiments, we decided to use both types of measures:
\begin{itemize}
    \item cosine similarity for embeddings with Hugging Face model: 'paraphrase-MiniLM-L6-v2'
    \item BERTScore employing DistilBERT, which returns an F1 score
\end{itemize}

Cosine similarity (\ref{eq:cos_sim}) is calculated as follows:
\begin{equation} \text{Cosine Similarity}(\mathbf{A}, \mathbf{B}) = \frac{\mathbf{A} \cdot \mathbf{B}}{\max(\|\mathbf{A}\|, \varepsilon) \cdot \max(\|\mathbf{B}\|, \varepsilon)} 
\label{eq:cos_sim}
\end{equation}
Where A and B are two vectors, specifically the embedding representations of the reference and generated text sentences, the max function and epsilon parameter are used to avoid division by zero or very small numbers.
\subsection{Classification Improvement}
We measure the improvement in classification by fine-tuning two models, LaBSE \cite{feng2022languageagnosticbertsentenceembedding}, and distilBERT \cite{sanh2020distilbertdistilledversionbert}, both showing impressive performance in multiple classification tasks, including emotion classification. Then, we compare the results with the performance of models before fine-tuning. We measure the change in the f-1 measure for both the entire dataset and the f1-macro measure for the augmented classes.

\section{Results}

\begin{table}[h!]
\centering
\adjustbox{max width = \textwidth}{
\begin{tabular}{l|l|l|l|l|l|l}
\textbf{Data aug.} & \textbf{Word} & \textbf{Word} & \textbf{Word} & \textbf{Jaccard} & \textbf{Entropy} & \textbf{TTR}\\
& \textbf{Original} & \textbf{Generated} & \textbf{Ratio} & \textbf{Dissimilarity} & & \textbf{Ratio} \\ \hline
\textbf{Prompt 1 GPT-3.5} &\textbf{12} & \textbf{16}&\textbf{1.4146}&\textbf{0.8190} &\textbf{1.1523}&\textbf{1.2167}\\
Prompt 2 GPT-3.5 Nmax &11&13&1.3148&0.8069&1.1250&1.1876\\
Prompt 2 GPT-3.5 Nbal &12&15&1.3900&0.8117&1.1410&1.2164\\
Prompt 1 GPT-4 & 12&14&1.2479&0.8229&1.1010&1.1733\\
\textbf{Prompt 2 GPT-4 Nmax} &\textbf{11}&\textbf{12}&\textbf{1.2419}&\textbf{0.8233}&\textbf{1.1055}&\textbf{1.1858}\\
Prompt 2 GPT-4 Nbal &12&14&1.2414&0.8184&1.1027&1.1862\\
\hline
\textbf{BT with MarianMT}&\textbf{12}&\textbf{14}&\textbf{1.9729}&\textbf{0.7992}&\textbf{0.9157}&\textbf{1.1748}\\
BT with DeepL&13&13&1.0093&0.5036&1.0006&1.0016\\
BT with gpt-3.5&13&14&1.1380&0.6152&1.0482&1.0554\\
BT with gpt-4&13&14&1.0861&0.5757&1.0303&1.0462\\
\textbf{BT with gpt-4-turbo}&\textbf{13}&\textbf{21}&\textbf{2.3324}&\textbf{0.9636}&\textbf{1.3250}&\textbf{1.4611}\\
\end{tabular}
}
\caption{Average results of Lexical Diversity for paraphrasing methods}
\label{tab:diversity}
\end{table}

\begin{table}[h!]
\centering
\begin{tabular}{l|l|l}
\textbf{Data aug.} & \textbf{Cosine Similarity} & \textbf{Bertscore-F1} \\ \hline
\textbf{Prompt 1 GPT-3.5} &\textbf{0.7031}&\textbf{0.8365}\\
Prompt 2 GPT-3.5 Nmax &0.6903&0.8324\\
Prompt 2 GPT-3.5 Nbal &0.6774&0.8299\\
Prompt 1 GPT-4 &0.6960&0.8461\\
Prompt 2 GPT-4 Nmax &0.6774&0.8354\\
Prompt 2 GPT-4 Nbal &0.6928&0.8400\\
\hline
BT with MarianMT&0.4758&0.7848\\
\textbf{BT with DeepL}&\textbf{0.8733}&\textbf{0.9248}\\
BT with gpt-3.5&0.8092&0.8989\\
BT with gpt-4&0.8465&0.9127\\
BT with gpt-4-turbo&0.1516&0.6892\\
\end{tabular}
\caption{Average results of Semantic Fidelity for paraphrasing methods}
\label{tab:fidelity}
\end{table}

\subsection{Lexical Diversity and Semantic Fidelity of generated data}
Looking at the diversity measures of the paraphrasing methods (Table~\ref{tab:diversity}), GPT-3.5 created longer sentences and had a higher TTR ratio than GPT-4. However, the Jaccard dissimilarity indicates that the GPT-4 methods were better at introducing lexical diversity. BT techniques (Table~\ref{tab:diversity}) introduced less diversity overall, except when using MarianMT models and the GPT-4-turbo, where the results indicated a high diversity. The best fidelity of meaning for paraphrasing (Table~\ref{tab:fidelity}) was achieved with prompt 1 for GPT-3.5-turbo and GPT-4, both for cosine similarity and BERTScore. BT techniques (Table~\ref{tab:fidelity}) received better results of generated text than paraphrasing methods, except for Hugging Face models and GPT-4-turbo. Using DeepL and GPT-4 presented the highest results.

\begin{table}[h!]
\centering
\begin{adjustbox}{width=\textwidth}
\begin{tabular}{c|c|c|c|c|c|c|c}
\textbf{Data aug} & \textbf{FT Model} & \textbf{F1-macro}  & \textbf{\%Change} & \textbf{F1-macro}  & \textbf{\%Change}& \textbf{F1-macro}  & \textbf{\%Change} \\
 && \textbf{(all Cls)} &\textbf{(all Cls)}  &\textbf{(aug Cls)}  & \textbf{(aug Cls)}  & \textbf{(othr Cls)}  & \textbf{(othr Cls)} \\ \hline
    \multirow{2}{*}{Baseline}
        & LaBSE & 0.467 & 0.00 & 0.211 & 0.00 & 0.522 & 0.00 \\
        & distilBERT & 0.458 & 0.00 & 0.174 & 0.00 & 0.520 & 0.00 \\
    \hline
    \multirow{2}{*}{Oversampling 3x}
        & LaBSE & 0.477 & 2.19 & 0.330 & 56.67 & 0.509 & -2.59 \\
        & distilBERT & 0.474 & 3.43 & 0.229 & 31.34 & 0.527 & 1.39 \\
    \multirow{2}{*}{\textbf{Oversampling 5x}}
        & \textbf{LaBSE} & \textbf{0.484} & \textbf{3.62} & \textbf{0.327} & \textbf{55.39} & \textbf{0.517} & \textbf{-0.92} \\
        & \textbf{distilBERT} & \textbf{0.481} & \textbf{5.06} & \textbf{0.307} & \textbf{76.12} & \textbf{0.519} & \textbf{-0.11} \\
    \hline
    \multirow{2}{*}{Prompt 1 GPT-3.5}
        & LaBSE & 0.473 & 1.35 & 0.276 & 30.94 & 0.516 & -1.25 \\
        & distilBERT & 0.472 & 3.10 & 0.286 & 64.06 & 0.513 & -1.34 \\
    \multirow{2}{*}{\textbf{Prompt 2 GPT-3.5 Nmax}}
        & \textbf{LaBSE} & \textbf{0.489} & \textbf{4.74} & \textbf{0.388} & \textbf{84.15} & \textbf{0.511} & \textbf{-2.23} \\
        & distilBERT & 0.474 & 3.56 & 0.266 & 52.81 & 0.520 & -0.03 \\
    \multirow{2}{*}{Prompt 2 GPT-3.5 Nbal}
        & LaBSE & 0.479 & 2.66 & 0.285 & 35.41 & 0.521 & -0.21 \\
        & distilBERT & 0.476 & 3.86 & 0.262 & 50.23 & 0.522 & 0.49 \\
    \multirow{2}{*}{\textbf{Prompt 1 GPT-4}}
        & LaBSE & 0.486 & 4.22 & 0.313 & 48.32 & 0.524 & 0.35 \\
        & \textbf{distilBERT} & \textbf{0.484} & \textbf{5.70} & \textbf{0.307} & \textbf{76.23} & \textbf{0.523} & \textbf{0.56} \\
    \multirow{2}{*}{Prompt 2 GPT-4 Nmax}
        & LaBSE & 0.478 & 2.36 & 0.276 & 30.80 & 0.522 & -0.14 \\
        & distilBERT & 0.466 & 1.77 & 0.221 & 26.69 & 0.520 & -0.05 \\
    \multirow{2}{*}{Prompt 2 GPT-4 Nbal}
        & LaBSE & 0.481 & 3.11 & 0.256 & 21.45 & 0.530 & 1.50 \\
        & distilBERT & 0.466 & 1.72 & 0.213 & 22.22 & 0.521 & 0.23 \\
    \hline
    \multirow{2}{*}{\textbf{0-shot gpt-3.5}}
        & \textbf{LaBSE} & \textbf{0.492} & \textbf{5.38} & \textbf{0.324} & \textbf{53.68} & \textbf{0.528} & \textbf{1.14} \\
        & distilBERT & 0.472 & 3.10 & 0.255 & 46.56 & 0.520 & -0.07 \\
    \multirow{2}{*}{0-shot gpt-4}
        & LaBSE & 0.474 & 1.61 & 0.243 & 15.09 & 0.524 & 0.42 \\
        & distilBERT & 0.482 & 5.17 & 0.280 & 60.79 & 0.526 & 1.12 \\
    \multirow{2}{*}{\textbf{5-shot gpt-3.5}}
        & LaBSE & 0.485 & 4.03 & 0.326 & 54.53 & 0.520 & -0.40 \\
        & \textbf{distilBERT} & \textbf{0.488} & \textbf{6.55} & \textbf{0.309} & \textbf{77.55} & \textbf{0.527} & \textbf{1.38} \\
    \multirow{2}{*}{5-shot gpt-4}
        & LaBSE & 0.489 & 4.72 & 0.310 & 47.08 & 0.527 & 1.00 \\
        & distilBERT & 0.479 & 4.50 & 0.289 & 66.07 & 0.520 & 0.01 \\
    \hline
    \multirow{2}{*}{\textbf{BT with DeepL}}
        & \textbf{LaBSE} & \textbf{0.497} & \textbf{6.47} & \textbf{0.377} & \textbf{78.74} & \textbf{0.523} & \textbf{0.13} \\
        & \textbf{distilBERT} & \textbf{0.494} & \textbf{7.77} & \textbf{0.387} & \textbf{121.99} & \textbf{0.517} & \textbf{-0.55} \\
    \multirow{2}{*}{HuggingFace models}
        & LaBSE & 0.490 & 5.10 & 0.340 & 61.22 & 0.523 & 0.18 \\
        & distilBERT & 0.473 & 3.27 & 0.298 & 71.30 & 0.511 & -1.68 \\
    \multirow{2}{*}{BT with gpt-3.5}
        & LaBSE & 0.487 & 4.41 & 0.293 & 38.82 & 0.530 & 1.40 \\
        & distilBERT & 0.486 & 5.66 & 0.348 & 99.83 & 0.516 & -0.79 \\
    \multirow{2}{*}{BT with gpt-4}
        & LaBSE & 0.493 & 4.39 & 0.341 & 62.03 & 0.526 & 0.71 \\
        & distilBERT & 0.483 & 5.46 & 0.313 & 79.39 & 52.02 & 0.07 \\  
    \multirow{2}{*}{BT with gpt-4turbo}
        &LaBSE&0.482&3.40  &0.309&46.99 & 0.520& -0.41\\
        & distilBERT & 0.479  & 4.75 & 0.313&79.97 & 0.516& -0.72\\
    \hline
\end{tabular}
\end{adjustbox}
\caption{Comparison of performance change for experimental setups.}
\label{tab:generating}
\end{table}

\subsection{Impact on Classification}
Looking at the results of fine-tuning LaBSE and distilBERT models on augmented datasets (Table~\ref{tab:generating}), there is a visible increase in performance with regard to augmented classes, as well as the entire dataset across all experimental setups. There is also no significant observable decrease in the performance of non-augmented classes, which means that improvement in the augmented classes does not happen at the cost of the rest of the dataset. Even baseline oversampling improved the model's ability to generalize. The mere addition of copies of the examples to the training set to increase results in a gain of more than 76\% with distilBERT in the augmented classes and 5\% on the entire dataset with negligible impact on the classification of other classes. Oversampling with more samples predictably yielded better results.

When it comes to paraphrasing, the smallest increase was approximately 1.3\%  and 30\% increase on the dataset and augmented classes with prompt 1 using the LaBSE model. However, the highest results were achieved with Prompt 1 GPT-4 with DistilBERT and Prompt 2 GPT-3.5 nmax with the LaBSE model, resulting in increases of more than 76\% and 84\%, respectively. In particular, Prompt 2 GPT-3.5 Nmax setup achieved the largest decrease in performance in other classes simultaneously, although it was only 2.23\%. The results with LaBSE were slightly better in all methods. Looking at the purely generative approach, the highest gain was reported for 5-shot GPT-3.5, showing a 6.5\% and 77.5\% increase with DistilBERT, and for 0-shot GPT-3.5 with LaBSE. These two experiments also both resulted in a slight increase in the performance of the non-augmented classes. No significant difference was between the results achieved by 0-shot and 5-shot learning.
BT achieved better results than the previous methods (Table~\ref{tab:generating}). The best gain was reported with backtranslation using DeepL with disilBERT, showing a 7. 7\% increase in the F1-macro in the entire data set and a greater 121\% increase in the augmented classes that decreased only by 0. 55\% in the other classes.

It is worth noticing that there is significant variance in results achieved by particular experimental setups for each method. All the methods compared to the oversampling yielded the results underperforming 3x oversampling on both models and the results beating 5x oversampling. Another interesting observation is that GPT-4 achieved the best results for a particular augmentation method in only one instance and only on the distilBERT model. This leads to the conclusion that while modern Language models can effectively leverage the traditional approaches to data augmentation, their overall performance largely depends on the models used for augmentation and specific configuration elements such as prompt and example selection.

\section{Limitations and Future Research}
Data augmentation using large language models seems effective in improving performance and mitigating class imbalance, as shown with the GoEmotions dataset for multi-label classification. The techniques presented in this research have the potential for a broader application across various datasets and problems, but further experiments need to be performed to confirm these findings. Future research and experiments should focus on:

\begin{itemize}
    \item methods using different models and datasets,
    \item experimenting with different parameters of the model,
    \item evaluation of generated text with LLM-as-a-judge,
    \item specific setups for backtranslation and multiple prompts for both paraphrasing and generation,
    \item evaluating the impact of LLM and in-context learning biases on the data augmentation,
    \item addressing cultural and global biases resulting from sourcing data from Reddit, e.g. by enriching data with other datasets,
\end{itemize}

\section{Conclusions}
We conducted experiments on multiple different approaches for data augmentation utilizing LLM. 
All methods produced semantically similar but distinct samples. Paraphrasing showed greater lexical diversity, while backtranslation maintained better semantic similarity, except for HuggingFace models and GPT-4-turbo. All methods improved the classification results. Although the improvements were modest, they are significant given the augmented classes' underrepresentation. The best F1-macro scores were achieved using the GPT-4 model for paraphrasing. Zero-shot and few-shot learning generally outperformed paraphrasing, with the best results from GPT-3.5. Backtranslation produced the best overall results, with the highest classification outcome using the DeepL model.

 The best F1 scores across all classes were achieved using backtranslation with LaBSE and DistilBERT via DeepL, GPT-4 on LaBSE, and zero-shot learning with GPT-3.5 turbo. These methods and setups achieved an F1-macro score above 0.49. Analyzing the results of classifying only augmented classes, the greatest gains were observed with backtranslation. The F1 score increased from 0.17 up to 0.38.
Assessing different methods and setups should include the costs of data augmentation. The main costs include the complexity of creating augmented data, API usage fees, the time required for fine-tuning the model on the augmented training set, and the computational resources needed. While DistilBERT produced slightly worse results than LaBSE, it required significantly less training time and computational resources, making it a more suitable model when resources are limited. Zero-shot learning and few-shot learning require fewer input tokens compared to paraphrasing and backtranslation. Oversampling slightly mitigated the class imbalance problem, but data scarcity remains. It is the least demanding method for acquiring new data in terms of complexity and time.

\section*{Acknowledgements}
Financed by: (1) the National Science Centre, Poland (2021/41/B/ST6/04471);
(2) CLARIN ERIC (2024–2026), funded by the Polish Minister of Science (agreement no. 2024/WK/01);
(3) CLARIN-PL, the European Regional Development Fund, FENG programme (FENG.02.04-IP.040004/24);
(4) statutory funds of the Department of Artificial Intelligence, Wroclaw Tech;
(5) the Polish Ministry of Education and Science (“International Projects Co-Funded” programme);
(6) the European Union, Horizon Europe (grant no. 101086321, OMINO);
(7) the EU project “DARIAH-PL”, under investment A2.4.1 of the National Recovery and Resilience Plan.
The views expressed are those of the authors and do not necessarily reflect those of the EU or the European Research Executive Agency.
%
%
%
\bibliographystyle{splncs04}
\bibliography{bibliography}

\end{document}